 \documentclass[sigconf]{acmart}

\AtBeginDocument{%
  \providecommand\BibTeX{{%
    \normalfont B\kern-0.5em{\scshape i\kern-0.25em b}\kern-0.8em\TeX}}}

\usepackage{multirow}

\setcopyright{acmcopyright}
\copyrightyear{2023}
\acmYear{2023}
\acmDOI{XXXXXXX.XXXXXXX}

\acmConference[FAccT '23]{Make sure to enter the correct conference title from your rights confirmation emai}{June, 2023}{Chicago, USA}
\acmBooktitle{FAccT '23: ACM Conference on Fairness, Accountability, and Transparency, June,  2023, Chicago, USA} 
\acmPrice{15.00}
\acmISBN{978-1-4503-XXXX-X/18/06}

\begin{document}


\title{Optimization's Neglected Normative Commitments}
\renewcommand{\shorttitle}{Optimization's Neglected Normative Commitments}


\author{Benjamin Laufer}
\email{bdl56@cornell.edu}
\affiliation{%
  \institution{Cornell Tech}
  \streetaddress{2 West Loop Rd}
  \city{New York}
  \state{NY}
  \country{USA}
  \postcode{10044}
}

\author{Thomas Krendl Gilbert}
\email{tg299@cornell.edu}
\affiliation{%
  \institution{Cornell Tech}
  \streetaddress{2 West Loop Rd}
  \city{New York}
  \state{NY}
  \country{USA}
  \postcode{10044}
}

\author{Helen Nissenbaum}
\email{hn288@cornell.edu}
\affiliation{%
  \institution{Cornell Tech}
  \streetaddress{2 West Loop Rd}
  \city{New York}
  \state{NY}
  \country{USA}
  \postcode{20044}
}


\begin{abstract}


Optimization is offered as an objective approach to resolving complex, real-world decisions involving uncertainty and conflicting interests. It drives business strategies as well as public policies and, increasingly, lies at the heart of sophisticated machine learning systems. A paradigm used to approach potentially high-stakes decisions, optimization relies on abstracting the real world to a set of \textit{decision(s)}, \textit{objective(s)} and \textit{constraint(s)}. Drawing from the modeling process and a range of actual cases, this paper describes the normative choices and assumptions that are necessarily part of using optimization. It then identifies six emergent problems that may be neglected: 1) {Misspecified values} can yield optimizations that omit certain imperatives altogether or incorporate them incorrectly as a constraint or as part of the objective, 2) {Problematic decision boundaries} can lead to faulty modularity assumptions and feedback loops, 3) {Failing to account for multiple agents}' divergent goals and decisions can lead to policies that serve only certain narrow interests, 4) {Mislabeling and mismeasurement} can introduce bias and imprecision, 5) {Faulty use of relaxation and approximation} methods, unaccompanied by formal characterizations and guarantees, can severely impede applicability, and 6) {Treating optimization as a justification} for action, without specifying the necessary contextual information, can lead to ethically dubious or faulty decisions. Suggestions are given to further understand and curb the harms that can arise when optimization is used wrongfully.


\end{abstract}



\maketitle


\section{Introduction} \label{sec:intro}

Mathematical optimization\footnote{For the formal definition used throughout this paper, see Appendix \ref{app:opt}.} is a way to relate a set of potential decisions to a specific goal and use this relationship to identify the best-performing decision. It is a paradigm used in industry and government and aids in developing a variety of computing tools including machine learning (ML) models. Optimization’s widespread use and applicability is enabled, in part, by its level of abstraction. By representing real decisions as a set of \textit{decision variable(s)}, \textit{objective(s)}, and \textit{constraint(s)}, optimization methods identify solutions that maximize the objective while meeting the relevant constraints. 

However, optimization does not necessarily imbue a decision with legitimacy or ethical justification. Because its parameters and functions are left to be specified, an optimization can be designed to serve subjective or parochial interests, out of step with social welfare. For example, when a social media advertiser uses optimization to target ads in a way that maximizes user engagement, it may not be the case that the so-called `optimized' decision is one that is best for society. 

This paper aims to highlight optimization’s normative choices and assumptions. In Section \ref{sec:lifecycle}, we describe the \textit{life cycle} of an optimization, particularly as it is used in operational decision-making, machine learning, and research. Our treatment of the life cycle or pipeline associated with applied optimization is intended to be reminiscent of existing characterizations of the \textit{machine learning life-cycle} which have proven useful for end-to-end analysis \cite{suresh2021framework,hutchinson2021towards,garcia2018context,cao2021toward}. We then focus on the various choices and assumptions inherent in the optimization life cycle, which we categorize into the following components: modeling, measurement, computing, and application. In Section \ref{sec:neglect}, we identify six emergent issues that may be neglected: \textit{1) misspecified values, 2) problematic decision boundaries, 3) failing to account for multiple agents, 4) mislabeling and mismeasurement, 5) faulty use of relaxation and approximation, and 6) treating optimization as a justification for action.} Finally, Section \ref{sec:programme} articulates suggestions and directions for research on optimization and its normative implications.



\section{The Optimization Life Cycle}\label{sec:lifecycle}

Examining the assumptions and choices laden in an optimization requires thinking about optimization as a sociotechnical process. In this section, we describe the typical steps in the \textit{life cycle} (or \textit{pipeline}) associated with applying optimization in real-world settings. An equivalent life cycle for ML systems has been developed and widely used (see e.g. \cite{souza2019provenance, garcia2018context}) including as a framework for evaluating normative assumptions and potential harms  \cite{suresh2021framework, lee2021risk, ashmore2021assuring, cao2021toward,lauferend}. We propose an analogous pipeline that describes the use of optimization. Equipped with such a pipeline, we will systematically analyze each of the component choices and assumptions in order to unearth normative commitments. A visual representation of the optimization life cycle is provided in Figure \ref{fig:lifecycle}.

Although our goal is to provide a general characterization of the use of optimization from its inception (measurement and modeling) to its execution (solving and applying), it is important to note that optimization can look very different depending on the context. Thus, we'll start by discussing some domains where optimization is commonly applied. Then, we'll discuss the component steps involved in a typical optimization process.

\subsection*{Common Contexts}

Here we discuss three common application domains for optimization: operational decision-making, machine learning, and scientific research.

\subsubsection*{Operational Decision-Making} \label{subsec:operational}

Optimization is perhaps best understood as a method to approach complex logistical decisions. These decisions aim for operational efficiency for large-scale industrial or institutional endeavors with significant complexity. Canonical examples cited in introductory optimization courses include signal processing and networked communication \cite{luo2003applications, neely2010stochastic}, portfolio optimization in financial mathematics \cite{black1992global, perold1984large, cvitanic1992convex, jorion1992portfolio}, and transportation \cite{bielli2002genetic, mandl1980evaluation, osorio2013simulation}. 

For examples of high-impact optimization problems applied to decision-making, one can simply look to the history of linear programming.\footnote{Linear programming is one particular type of optimization where the constraints are a set of linear inequalities and the objective is a linear equation. In the first half of the twentieth century, optimization was more commonly known as `mathematical programming,' where the word \textit{programming} referred to logistical decision-making rather than computer programs.} LP problems gained prominence in both the US and USSR during the Second World War, where mathematicians and economists in both countries began considering how to devise optimal large-scale policies with limited budgets --- these policies ranged from wood production to air force allocation \cite{singhbrief}.

\subsubsection*{Machine Learning} \label{subsec:ml}

Optimization and machine learning are intimately related and machine learning often relies on optimization. The algorithms used to train a ML model may use a formal notion of \textit{loss}---for example, prediction error---in the training process. Fitting parameters to \textit{minimize} this loss can be written as an optimization problem.

For example, a univariate linear regression using ordinary least squares (OLS) finds a line to fit the data. Of all possible lines which could be used to describe the relationship in the data, the `line of best fit’ according to OLS is the line which minimizes the sum of squared errors between the predicted values and the observed data. This task can be written as an optimization problem, where the decision variables are the slope $m$ and the intercept $b$ of the line. The feasible set is all real numbers (the problem is unconstrained beyond $m,b \in \mathbb{R}^2$). And the objective is to minimize loss, namely the sum of squared errors.

For more complex models using high-dimensional datasets, the optimizations become more tedious to write out but the idea is the same: Some notion of loss is defined so that the model fit will resemble the observed training data, and defining the model involves solving an optimization to minimize loss.

Although we predominantly draw from examples in logistical decision-making and management science, we believe that this paper’s framework yields useful insights about optimization’s role in the machine learning pipeline. For example, defining loss is a choice with particular priorities and commitments. Sometimes, this decision is made simply by choosing a loss metric which is computationally easy to work with. Another example of this paper’s usefulness in scrutinizing ML is in assumptions about decisions and decision boundaries: Often, ML predictions are translated into decisions without considering the impacts, social implications, or feedback effects associated with the particular decision.


\subsubsection*{Scientific Research} \label{subsec:sci}

Optimization crops up in a number of disciplines outside of operations research, math and statistics. Where optimization is applied to scientific research, it may guide design decisions for building scientific instruments \cite{chen2005design, niu2014design, zheng2012design} or in the discovery of new drugs to test \cite{ou2012computational, prada2016application, caldwell2001new, de2020silico, copeland2013evaluation}. These are examples of \textit{operational} uses, similar to those described above.

Another use for optimization in research is as a descriptive tool. For example, researchers employ agent-based models to understand complex processes, and optimization often plays a role in simulating how modeled agents behave. The use of optimization in simulation or theoretical models is distinct because it is used to describe the behavior of an agent. The optimization is used to draw descriptive conclusions that do not necessarily guide real-world decisions. Optimization in this sense \textit{describes} rather than \textit{dictates}. For example, in game theory’s applications to economics \cite{sohrabi2020survey, aghassi2006robust,curiel2013cooperative, scutari2010convex} and biology \cite{hammerstein1994game, dugatkin2000game, grupen2020low, leboucher2018enhanced}, optimization is used to model agent decisions and strategies. Pinpointing assumptions and normative imperatives within these optimizations may be useful---they may reveal values and priorities held by researchers, or assumptions made about economic subjects. However, because these optimizations are not used to justify a particular `optimized' decision, we do not draw examples from these domains and consider them to be out of scope.

\begin{figure*}
    \begin{center}
    \includegraphics[width = \textwidth]{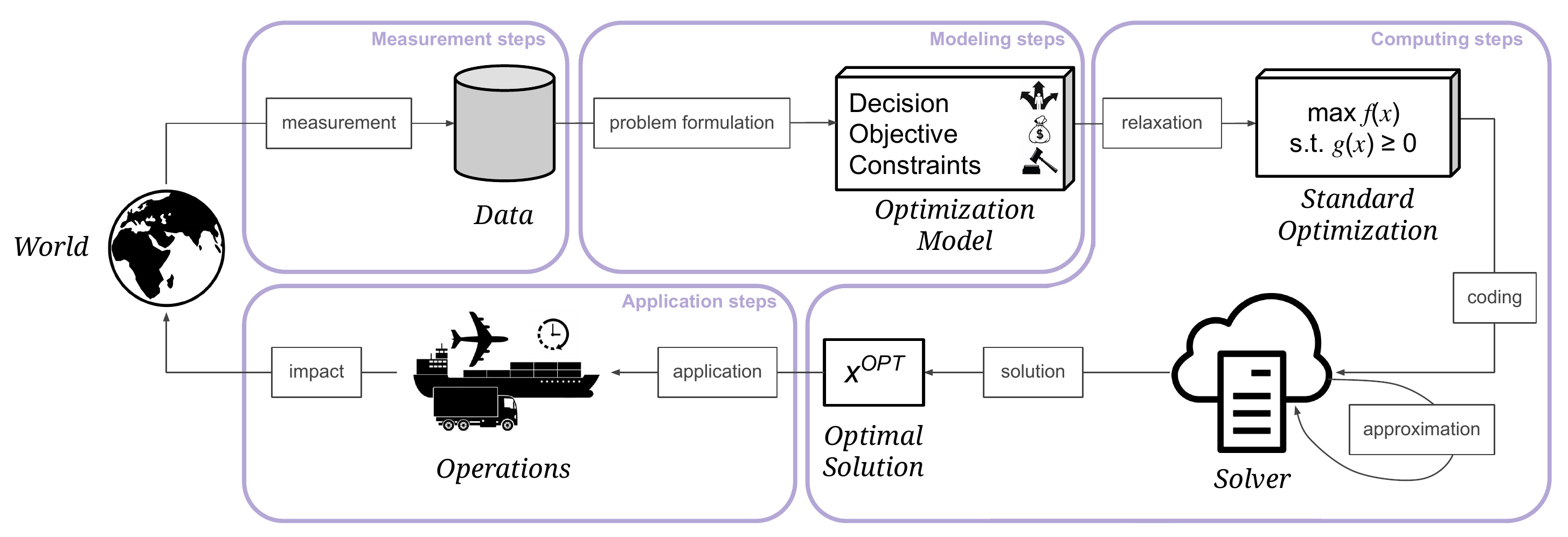}
    \end{center}
    \caption{Optimization life cycle, beginning with measurement and data generation and ending with operational decision-making and real-world impact.}
    \label{fig:lifecycle}
    
\end{figure*}



\subsection*{Choices, Abstractions and Assumptions}\label{sec:assumptions}

Formulating an optimization problem requires transforming complex information in the world into a simplified model. Models of any sort can (ideally) use abstraction and representation to make information communicable, actionable or useful \cite[p 20]{o2016weapons}. Therefore, models that are useful often omit or collapse a significant amount of information.

The idea of abstraction is familiar to computer scientists, whose introductory computer science classes often involve creating a simple \textit{class} or \textit{object} such as a `Bank Account' or an `Employee' (see e.g. \cite[p 316-319]{sommerville_2004}). Doing so does not involve building a physical bank, of course; instead students simply define a set of variables (such as name and salary) and functions (such as ``retire” or ``promote”). By removing significant material and contextual aspects of a real-world system, a computer program can simulate a few key functions in order to make a particular point or perform a particular task. Abstraction is powerful and crucial to the field of computing; however, it also poses challenges: through making assumptions and collapsing information, models risk misrepresenting or omitting key information.

In the following sections, we trace the component steps that constitute the optimization process. The process spans modeling, measurement, solving, and deploying, as shown in Figure \ref{fig:lifecycle}. Reasoning about these components requires assumptions--some axiomatic, others readily contestable. By systematically enumerating and formalizing these as constitutive parts of the optimization life cycle, the system's risks, harms, indeterminacies,\footnote{See \citet{dobbe2021hard}, which similarly uses assumptions, indeterminacies and \textit{vague} aspects of socio-technical systems to explore their normative implications.} and concealed normative positions can be more easily identified and communicated.\footnote{We do not claim to provide an exhaustive method for anticipating future harms or \textit{verifying} optimizations. Our project is influenced by \citet{smith1985limits} on this matter: we seek to disambiguate and excavate the social and political aspects of optimization as a method, rather than make grand claims about its use.}

\subsection{Modeling} \label{subsec:model}

A preliminary step in putting forward an optimization model is defining the set of variables and parameters that will be included in the model. These variables are then used to formalize the objective function, decision variable and constraints which together define the feasible set of potential decisions and their corresponding utility or performance. As the model is defined and worked with, modeling choices play a determining role in the larger optimization process and any resulting decisions.

In this section, we focus on the three components of a standard optimization and discuss the negotiable decisions and assumptions at play during the modeling process. We refer to Appendix \ref{app:opt} for the standard form of an optimization problem.

\subsubsection*{The Objective Function}  Specifying an objective function, utility function or goal immediately invokes values judgements. Formalizing an objective function often means relying on quantifiable notions of utility or welfare that are relevant for a given decision. Formalization of this sort can also involve omitting values or desiderata from consideration. When multiple goals are valuable to consider, modeling assumptions are needed to incorporate these goals into a single objective. Often, this process involves defining a \textit{welfare function}, a term borrowed from economics (see e.g. \cite{kaneko1979nash, coleman1966possibility}). Instantiating an objective of this sort typically involves weighing the relative importance of different goals, and characterizing the overall objective function’s mathematical properties, like whether it is convex \cite{boyd2004convex}. 

\subsubsection*{The Decision Variable} Specifying the decision variable(s) requires making a number of assumptions about the decision-making agent and their degree of control. The optimization paradigm presumes that a decision-maker has a certain set of options, and inquiry is narrowed to only comparing between the options. Such an approach avoids thorny questions about who is in control and whether they ought to be. Consider, for example, an optimization aimed at setting a regulatory fine. Defining the decision mathematically may necessarily involve stripping away particularities about how fines result from complex bureaucratic and deliberative processes. Instead, an optimization might model the decision simply as a single variable representing the dollar value of a fine. Abstractions of this sort can appear self-evident even as they conceal choices with decisive normative commitments.

\subsubsection*{The Constraints} Specifying constraints is also known as specifying the \textit{feasible set} of decisions. This step involves making assumptions about the physical relationship between variables. For instance, an optimization problem may try to maximize production subject to a budgetary constraint. Budgets often play into optimizations and models must specify what kind of scarcity exists and how it is being measured: is the agent constrained monetarily? spatially? temporally? Though the feasible decision space is traditionally conceived of as the set of possible decisions, constraints are often used to guarantee that a decision meets some baseline ethical or social standard. We call these sorts of constraints \textit{values-driven constraints}, and note that \textit{fairness} is a particularly common values-driven desideratum that gets modeled as a constraint.\footnote{See \citet{passi2019problem} on problem formulation and modeling in the context of fairness.} Section \ref{subsubsec:mistakenlymodelingconstraint} includes a more in-depth discussion of the modeling assumptions involved in incorporating values into an optimization's constraints.

\subsection{Operationalization, Measurement and Quantification} \label{subsec:measure}

Applying optimization requires a number of decisions aimed at \textit{relating an abstract quantity to the external world}. Where modeling (discussed in Section \ref{subsec:model}) involves modes of knowledge aimed at making sense of the chaos of the world, measurement is how we empirically find new information or physically instrument the world. There is significant work on the elusive nature of measurement---here we highlight theoretical groundwork \cite{barad2003posthumanist, joslyn2001semiotics, finkelstein1984review, jacobs2021measurement, moss2021objective} as well as technical work on bias in measurement \cite{obermeyer2019dissecting, millsap1993methodology, hernan2009invited, bhatt2021uncertainty}. In this section we categorize optimization's measurement assumptions into two categories: 1) defining what is being measured and 2) deciding how to measure it.


\textit{What is being measured?} Deciding what is relevant or worthwhile to measure is an impactful step in setting up an optimization model. When written out on paper, an optimization has finite number of variables, and the definitions of these variables are often formally listed. These variables constitute a highly abstracted and stylized version of the world---few of their characteristics may correspond to the real-world entity that they try to mimic or count. For example, in public transit optimization, a bus might get modeled as a capacity, route, and fuel level---all other characteristics are removed from consideration.

\textit{How is it being measured?} Research on optimization often focuses on methods for solving and approximating certain types or groups of optimization problems. Accordingly, much of the attention devoted to optimization isn't concerned with the measurement of particular quantities. Instead, it is assumed that quantities are well-defined and measurable. Therefore, measurement is notably far afield from academic research related to optimization. When optimization is used to guide real-world decision-making, however, measurement becomes a necessary component to understand the impacts and constraints of a decision. Measurement is never perfect; and any measurement technique inevitably introduces \textit{error}. Designers of optimization models must decide what to do about error: for example, they may ignore it, leaving it out of the optimization, or they may use stochastic optimization \cite{heyman2004stochastic, sahinidis2004optimization} to directly model uncertainty.

\textit{Example: \textbf{Transportation}}.
To better understand how measurement considerations play into an optimization model, consider the example of a logistics company that aims to optimize routes for its delivery vehicles. To perform this sort of optimization, the company may want to consider the way that refueling constrains its decision. The resulting optimization may assume that trucks have a certain unchanging fuel efficiency in miles per gallon (MPG) to make it easy to relate route distances to gasoline use. Doing so would rely on a certain kind of measurement (average MPG for a vehicle) to relate two modeling variables (route distance and gas usage), without requiring further measurement about how fuel efficiency varies depending on truck speed, number of traffic lights, fuel grade, price, and other factors. Assuming a static fuel efficiency greatly reduces what must be measured by making a simplifying assumption about how a quantity is measured. 

\subsection{Computing} \label{subsec:compute}

A dominant strand of research on optimization is concerned with computing and computability. Research findings in the field of Operations Research include algorithms to solve optimizations (e.g. \cite{nelder1965simplex, johansson2008subgradient, beck2003mirror}) and methods to approximate optimal values (e.g. \cite{kempe2003maximizing, nikolova2010approximation, kasperski2006approximation}). In some cases, computability is in tension with accurately and comprehensively modeling the state of the world. For example, an objective function may be multi-modal or non-differentiable in practice, but accurately modeling these features would make the optimization computationally difficult to solve. As such, optimization commonly relies on \textit{relaxation} to transform a domain-specific, computationally-intensive optimization into a familiar, solvable type of optimization, ideally one that yields the same or a provably nearby solution. In the process of solving these optimizations, sometimes an optimal solution is still too computationally expensive to attain, so \textit{approximation algorithms} \cite{williamson2011design} are similarly used to find a solution that is nearly-optimal. In this section, we describe these two steps --- \textit{relaxation} and \textit{approximation} --- and aim to excavate their normative assumptions.

\subsubsection*{Relaxations} Relaxations are the transformation of an optimization into another optimization that is easier to solve or work with. The formal definition of the sort of relaxation we refer to in this paper is as follows:

\begin{definition}[Relaxation] \label{def: relax}
    A minimization problem $Q$ is said to be a relaxation of a minimization problem $P$ if the following conditions are met: \begin{enumerate}
        \item $F(P) \subseteq F(Q)$, where $F(x)$  is the feasible decision set defined by the decision and constraints of optimization $x$.
        \item The objective function of optimization $Q$ is less than or equal to that of $p$ for all parameter values in $F(P)$.  \cite{geoffrion1974lagrangean}
    \end{enumerate}
\end{definition}


\subsubsection*{Approximations} Approximations are quantities or methods that are nearly correct. Relaxing an optimization and using the solution as an answer to the original problem (as described above) is one type of approximation. Here we focus specifically on \textit{approximation algorithms} or ways of approximating certain types of optimization problems \cite{williamson2011design, vazirani2001approximation}.

\begin{definition}[Approximation algorithm] \label{def: approximation}
    An $\alpha$-approximation algorithm for an optimization problem is a polynomial-time algorithm that for all instances of the problem produces a solution whose value is within a factor $\alpha$ of the value of an optimal solution \cite[p 14]{williamson2011design}.
\end{definition}

Approximations can happen after an optimization has already been relaxed or fully defined; they do not necessarily rely on expanding the feasible set or changing the objective function as relaxations do. Instead, approximations are often algorithmic approaches that can be incorporated into a computational solver (see Figure \ref{fig:lifecycle}). Using approximations to solve an optimization problem requires making a number of assumptions---for example, it is assumed that a sub-optimal solution with nearly-optimal utility performance is acceptable as a decision outcome. Without formally reporting and considering performance guarantees ($\alpha$ in Definition \ref{def: approximation}), those who apply optimization methods may land on approximate solutions that are far from optimal without knowing it. Such pitfalls are discussed further in Section \ref{subsec:sloppy}.

Relaxations and approximations suggest a trade-off between at least two desiderata: reducing computational complexity and maintaining performance (e.g., $\alpha$). Of course, modelers are juggling a number of other potential goals as well, including consistency \cite{evers2009regrouping, fedrizzi2007incomplete,cooper2023variance}, robustness \cite{beyer2007robust, gabrel2014recent, ben2002robust, bertsimas2011theory}, and correctness \cite{samet1978proving, fluckiger2017correctness, d2015correctness}. Trade-offs between goals abound in optimization, and have been discussed especially in machine learning contexts \cite{belkin2019reconciling, zhang2019theoretically, kleinberg2016inherent, cooper2021emergent}. These trade-offs may or may not receive explicit consideration from model designers. However, whether or not the designer has explicitly taken these values into consideration or weighed between them, the underlying values and normative commitments are reflected in an optimized decision. 

\subsection{Application} \label{subsec:apply}

Optimization research tends to categorize and solve problems based on particular groupings or structures of problems. Often, solution algorithms are found through a process of typifying the the objective function and constraints and demonstrating solutions that work on certain general categories of problems (e.g., convex problems \cite{boyd2004convex}). Solutions, approximations and relaxations are then developed for these various already-abstract problems without considering domain-specific contexts or scenarios.

Our characterization of optimization research as abstracted from real-world scenarios is not intended as an indictment of the research but instead provides lessons for how optimization might be safely and ethically applied in real-world domains. When applying optimization, it is essential to consider whether the application is reasonable in the real, non-abstract context, or whether it is is, for example, far-fetched, impossible, or unethical. Especially when objectives are contested, numerous, complex or conflicting, researchers and practitioners sometimes find that the right choice is refusal to adopt or use a technological tool \cite{simpson2007ethnographic, zajko2022artificial}.

When optimization is applied in real-world settings, decision-makers necessarily make assumptions about \textit{applicability} to conclude that an optimal solution is the right course of action. Assumptions are made about both the \textit{impact} of the optimization
and the \textit{relation} between the decision and other decisions. Often, even though individual decisions are said to be optimized, the process of optimization requires tweaking, iteration, and repeated use over multiple decisions.\footnote{Failing to model the decision as dynamic has been noted by \citet{selbst2019fairness} as a potential vulnerability for decision-making in socio-technical systems. The authors refer to this issue as the  \textit{ripple effect trap}.} The dynamics arising from  applicability assumptions may generate forms of instability: system oscillations \cite{zhu2018optimization}, runaway feedback effects \cite{ensign2018runaway, lum2016predict}, amplification, or domain drift \cite{tomani2021post} over time.




\section{Emergent Problems and Neglect}\label{sec:neglect}

The optimization paradigm requires a number of choices, simplifications, and assumptions, as demonstrated in Section \ref{sec:assumptions}. These choices can contain particular commitments and serve particular interests. A company's goals, priorities, values, norms, and conventions can influence the considerations that end up in an optimization model. As such, `optimal' decision-making in one company or sector means something particular and context-dependent. A decision described as `optimal' or `optimized' is not, necessarily, a virtuous decision.

Our contention is that optimization is a powerful tool; its use is not ethically neutral simply because it is quantitative and scientific.\footnote{Relatedly, applications of \textit{mechanism design} have been referred to as `technologies of depoliticization' \cite{hitzig2020normative}. See also \citet{viljoen2021design}, \citet{hitzig2019technological}, and \citet{finocchiaro2021bridging} on mechanism design’s normative dimensions. Mechanism design can be described as a type of optimization and, similarly, can invoke objectivity or impartiality.} Optimization can be used for a number of desirable or undesirable ends, and its application can lead to emergent issues and downstream consequences. We do not argue that optimization’s constituent choices and assumptions are inevitably problematic. Instead, we use the process (or life cycle) of optimization to identify ethical pitfalls, harms, and wrongs that can emerge. 

\subsubsection*{When are these problems `neglected'?} To neglect is to \textit{fail to care for properly}. The worry with optimization is that it can appear to be objective and exhaustive. Optimization is a systematic paradigm used to approach decisions, constraints, and goals, however, its use does not guarantee that all relevant decisions, constraints, goals, interests, values, and other considerations have been adequately weighed. As we describe in this section, there are several cases where optimizations do not properly account for particular harms or values. When an optimization does not take proper consideration of ethically relevant interests or consequences, these interests or consequences may be described as \textit{neglected}.

To be sure, some emergent problems and downstream consequences are unknowable and unforeseeable. For example, consequences that are far enough in the future, or that arise from apparent anomalies, may be exceedingly difficult to predict and account for in a decision calculation. These harms and wrongs are difficult to attribute to carelessness or malice. But, by observing and documenting known problems and failure modes, these cases can inform standards of care and accountability mechanisms.

This section puts forward six emergent issues that can be overlooked when optimization is used to justify or derive a decision. These issues arise at various steps throughout the optimization life cycle. Where relevant, we provide examples and offer suggestions.

\begin{table*}[htp]
\caption{Life cycle components, categories of neglected problems, and particular emergent issues in optimization.}
\label{table:summary}
\footnotesize
\begin{center}
\begin{tabular}{l l l}
\toprule
\textbf{Life Cycle Steps} &   \textbf{Emergent Problems}   &   \textbf{Particular Issues}   \\
\midrule
\multirow{7}[7]{*}{Modeling (\ref{subsec:model})}   

   &   \multirow{3}[3]{*}{Misspecifying Values (\ref{subsec:misspecifyingvalues})} & Omitting Values (\ref{subsubsec:omittingvalues}) \\
   
   \cmidrule{3-3}
   
   & & Improperly Modeling a Value as a Constraint (\ref{subsubsec:mistakenlymodelingconstraint}) \\
   
   \cmidrule{3-3}
   
   & & Improperly Modeling a Value as an Objective (\ref{subsubsec:mistakenlymodelingobjective})\\

   \cmidrule{2-3}
   
   & \multirow{2}[2]{*}{Problematic Decision Boundaries (\ref{subsec:imprecisedecisionboundaries})} &
            Faulty Modularity Assumptions (\ref{subsubsec:faultymodularity})  \\
    \cmidrule{3-3}
    &  & Feedback Loops (\ref{subsubsec:feedbackloops}) \\
        
\cmidrule{2-3}

   &   \multirow{2}[2]{*}{Neglecting to Consider Multiple Agents  (\ref{subsec:neglectingtoconsidermultipleagents})} & Objective Reflects Narrow Interests (\ref{subsubsec:objectiverepresentsnarrowinterests})  \\
   
   \cmidrule{3-3}
   
   & & Decision Model Formalizes Notions of Control (\ref{subsubsec:decisionmodelformalizes})\\

   \midrule
\multirow{2}[2]{*}{Measurement (\ref{subsec:measure})}   &   \multirow{2}[2]{*}{Mislabeling and Mismeasurement (\ref{subsec:labelmismeasurement})} &
            {Imprecise Labels and Measurements (\ref{subsubsec:impreciselabels})}      \\
    \cmidrule{3-3}
    & & Biased Labels and Measurements (\ref{subsubsec:biasedlabels}) \\

\midrule
\multirow{2}[2]{*}{Computing (\ref{subsec:compute})}   &   \multirow{2}[2]{*}{Faulty Solution Steps (\ref{subsec:sloppy})} & Faulty Use of Relaxations (\ref{subsubsec:sloppyrelaxations})
                \\
    \cmidrule{3-3}
    & & Faulty Use of Approximation Algorithms (\ref{subsubsec:sloppyapproximation})  \\

\midrule
\multirow{2}[2]{*}{Application (\ref{subsec:apply})}   &   \multirow{2}[2]{*}{Using ``Optimal'' as Justification (\ref{subsec:usingoptimalasjustification})} & Justificatory Language (\ref{subsubsec:language})\\
    \cmidrule{3-3}
    & & Reverse Engineering/Retrofitting (\ref{subsubsec:formal})  
            \\
   
\bottomrule
\end{tabular}
\end{center}
\end{table*}

\subsection{Misspecifying Values} \label{subsec:misspecifyingvalues}

Optimization requires formally specifying at least one decision and at least one measure of utility. The values laden in optimization are made most apparent by the existence of an objective function, which formalizes the goal that the decision-maker is assumed to wish to maximize. Aligning optimization objectives with social values is a topic that has received significant interest \cite{hadfield2016cooperative, milli2021optimizing}. But values play into other components of a model besides the objective. For instance, models can include constraints motivated by ethical norms or values like fairness.

Here, we discuss the issues that arise when an optimization model is misspecified such that values are not given proper consideration. We identify two mechanisms that result in misspecified values: 1) leaving values out of an optimization completely and 2) mistreating or mismodeling values within an optimization. In discussing the latter type of misspecification, we consider instances where values are wrongfully incorporated as optimization constraints (instead of objectives) or as objectives (instead of constraints). 

\subsubsection{Omitting Values} \label{subsubsec:omittingvalues}

Omitting values from an optimization problem can be either careless or deliberate. A careless omission of values arises when the humans designing and executing an optimization fail to properly consider the ramifications of their choices. A deliberate omission of values arises when the humans do consider the ramifications of their choices, and then purposefully choose to ignore or conceal them. For example, a company that makes custom shirts might decide where to source their textiles. An optimization problem that minimizes the cost of textiles might suggest changing which factory they contract with. However, it could be that the new factory has dangerous working conditions or otherwise harsh and unethical labor practices. If the optimization designer were unaware of this distinction, the omission of labor considerations from the decision would be \textit{careless}; if they were aware, it would be \textit{deliberate} and malicious.

Scrutinizing which values and implications are included or excluded from an optimization provides important information for attributing harms and establishing \textit{accountability}. However, the optimization model alone does not contain all the relevant and necessary information. For example, one cannot differentiate negligent and deliberate omissions simply by looking at an optimization problem---further contextual evidence is needed to qualify the human decisions behind the model. If somebody omits a value from an optimization deliberately, they may be using the optimization as a way of \textit{justifying} an otherwise faulty or problematic decision. We turn back to this idea in Section \ref{subsec:just}.

\subsubsection{Improperly Modeling a Value as a Constraint} \label{subsubsec:mistakenlymodelingconstraint}

An optimization model may include a constraint that is intended to operationalize a moral or ethical imperative. For example, a new housing development may be required to include a certain number of low-income or rent-stabilized units, leading developers to treat this requirement as a constraint on building a profitable new tower. As another example, when companies boycott sales in regions to voice dissatisfaction with foreign states’ political decisions, their resulting business strategy is constrained by their stance; even while their goal is profit and expansion. 

In the optimization paradigm, constraining a decision removes a set of choices from consideration. Constraints are categorical rules that determine the set of feasible choices. They are inflexible and not negotiable. Incorporating a values-driven constraint, therefore, lends itself to ethical considerations that are categorically satisfied over a range of (permissible) behaviors. Consider, as a toy example, a man who decides to go an a diet that forbids foods with added sugar. This diet \textit{constrains} the set of possible decisions the man can make. As long as he is not eating food with added sugars, he is free to eat whatever he pleases. The man's underlying goals are unknown to us: he might want to lose weight; he might want to snack less often; or he might want to protest the sugar industry. A constraint is categorically either satisfied or not satisfied---as long as it is satisfied, it does not guide a decision.

Though constraints play an important role in determining the feasible set of decisions available to a decision-maker, they can also be weak as ethics interventions. As long as two potential decisions are both allowed, constraints provide no reason to prefer one potential decision over the other. Put another way, within the feasible set, a constraint has no bearing on what option gets chosen. A constraint is described as \textit{trivial} if instituting it does not change the optimal solution. Even when a constraint is non-trivial, it tends to yield an optimal solution that lies on the constraint boundary: {\textit{solutions tend to `bump up against' constraints.}} For example, if a for-profit housing development must contain a certain minimum number of rent-stabilized units, an `optimal' business strategy might suggest including the minimum allowable rent-stabilized units, in order to maximize profit. In practice, the `optimal' solution might do the \textit{bare minimum} at meeting some values-driven imperative, if that imperative is modeled as a constraint. 

In computing contexts, values may be operationalized as a constraint or as an objective. For example, the value of \textit{{fairness}} might be operationalized as a constraint on an ML system, or alternatively, the ML system may be designed to advance \textit{{equity}} as its objective. These choices can profoundly impact the `optimal' decision.


\textit{Example: \textbf{Access}}. Consider a city which has two potential ways of providing transportation to its residents. One option is to allow private industry to provide transportation through contracts with the city. In this case, providers aim to \textit{maximize profit}, subject to certain \textit{contractual and legal constraints}. For instance, these companies could agree to provide a certain level of access to low-income neighborhoods. 
An alternative option is to treat transportation as a public service, where a single entity with a budget is tasked with providing transportation. In this public approach, an agency might aim to \textit{maximize access} subject to \textit{budgetary constraints}. These two approaches to optimization logic encode the same values (access and budgetary responsibility) but do so in ways that can lead to vastly different levels of access across communities.

\subsubsection{Improperly Modeling a Value as an Objective} \label{subsubsec:mistakenlymodelingobjective}
An optimization's objective is a way of defining utility as a quantifiable and measurable function of the decision variable. This way, for any two decision values, the utility of one decision can be quantitatively compared to the utility of the other decision. Defining the objective lends itself well to \textit{ends-based} or \textit{utilitarian} frameworks which treat value as quantifiable and to some extent predictable \cite{sinnott2003consequentialism}. 
An optimization model presumes to model the relationship between a decision and its utility outcome---in non-stochastic optimizations, this relationship is presumed to be fully known and defined, whereas stochastic optimization models randomness explicitly \cite{sahinidis2004optimization}. Sometimes, values are treated as objectives when they ought to be modeled elsewhere in an optimization. 

\textit{Example: \textbf{Safety}}. Car companies communicate that they aim to build the safest cars. In reality, these companies have many goals including engine efficiency, speed, comfort, and profit. If safety were truly the utmost objective, a car would not move: staying parked is the safest state for a car. Of course, the functionality of a car requires that safety is not always modeled as a \textit{sole} objective (even if such an objective is communicated in advertisements). More realistically, many car companies' safety considerations may be treated as \textit{constraints} on business decisions.


\subsection{Problematic Decision Boundaries} \label{subsec:imprecisedecisionboundaries}

A first step in specifying an optimization model is defining and constraining the \textit{decision variable}. Specifying which variables are freely tunable and which values are feasible lays the roadmap for solving the optimization. Doing so dictates the set of potential outcomes, how the problem may be solved, and, perhaps most importantly, which factors are within (versus outside) the control of the decision-maker.

Specifying the decision involves assumptions about \textit{who makes the decision} as well as \textit{what power they have over the state of the world}. Making such assumptions risks glossing over the mechanism through which a decision-maker enacts change, and a variety of requisite or related decisions that may be involved. The remainder of this section is devoted to 1) neglectful decision boundaries for synchronous decisions (which arise because of \textit{faulty modularity assumptions}) and 2) neglectful decision boundaries for asynchronous decisions (which can lead to \textit{feedback loops}).

\subsubsection{Faulty Modularity Assumptions} \label{subsubsec:faultymodularity}

It can be simpler or convenient to treat problems as separate optimizations when the number of decisions is too vast. The optimization paradigm makes it very easy to divide and conquer by treating the solution of one optimization as an (unchanging) parameter in a second problem.

Imprecise decision boundaries, including incorrect assumptions about modularity, lead to narrow optimizations that may not serve the broader function of a system or a society. Divide-and-conquer optimization strategies, with incorrect modularity assumptions, can have unanticipated normative and political implications. \citet[p 245]{mulligan2020concept} provide the example of access control to illustrate this point: although tech companies can justify switching from passwords to fingerprints to face ID by saying each new technology more \textit{optimally} performs the specified function of access control, such changes have notable political implications. For instance, each successive handoff might be said to curtail user control and diminish transparency.\footnote{See also \citet{goldenfein2020through}.} Improvements or optimizations directed at sub-functions of a larger system should consider broader societal implications and interests. Otherwise, optimization alone does not justify exchanging or updating a technological sub-component.

\textit{Example: \textbf{Transit}}. Consider a metropolitan transit authority faced with the following decisions: given budget and staff constraints, how much to charge in tolls and whether (or which) lanes should be restricted to high-occupancy vehicles during peak travel times. In actuality, these decisions are not independent: how much a highway charges in tolls will impact the number of cars on the road by inducing demand, which in turn is important for evaluating to what extent high-occupancy vehicle lanes will improve overall traffic throughput. Similarly, how many lanes will be restricted---or the window of time during which the restriction is enforced, or how costly the fine---will constrain how many people commit to carpooling and significantly affect budget and revenue considerations for the toll decision. These choices could be treated in isolation: first, decide how many restricted lanes would serve the carpooling population, and then (conditional on a set number of lanes) decide what to charge in tolls. But to do so would ignore the fundamentally interdependent nature of the decision, and may generate spillover effects that the simplistic model does not account for.

\subsubsection{Feedback Loops} \label{subsubsec:feedbackloops}

Optimizations may arise in a repeated fashion where it is unreasonable to assume each sequential decision is independent. For the same reason numerous synchronous decisions may be haphazardly split into separate optimizations (discussed in Section \ref{subsubsec:faultymodularity}), modelers may unreasonably assume sequential decisions are independent.

The phenomenon of feedback in optimization is represented in Figure \ref{fig:lifecycle}'s depiction of the entire optimization life cycle as a loop. Optimizations can be used repeatedly, even if the formal model does not account for one optimization’s impact on the next. Similar feedback loops have been observed in systems that do not necessarily involve optimization --- in domains ranging from mechanisms and acoustics to economic inequality, the concept of feedback and path-dependence explains patterns where compounding or overlapping impacts from multiple sequential events lead to unanticipated or chaotic outcomes. In the context of fairness for socio-technical systems, \citet[p 62]{selbst2019fairness} described this phenomenon (and corresponding neglect on behalf of designers) as the \textit{ripple effect trap}. In the case of optimization, likewise, it may be that an optimal decision in one instance is sub-optimal or actively harmful when repeated sequentially over time.

 
\textit{Example: \textbf{Predictive Policing}}. 
Consider the example of PredPol (now re-branded as Geolitica), a commercial predictive policing software that used historical data to statistically train `optimal'\footnote{The predictive algorithm developed by PredPol used expectation-maximization techniques developed in \cite{mohler2014marked}, as noted in \cite{mohler2015randomized} and \cite{ensign2018runaway}.} policing recommendations \cite{mohler2015randomized}. The software was accused of formalizing and amplifying historical biases \cite{lum2016predict, isaac2017hope, silva2018algorithms}. Subsequent studies used models of sequential predictive decisions to explain the reasons for these emergent biases: Decisions made as single-shot optimizations can introduce \textit{runaway feedback effects} when applied sequentially over time, since predictive algorithms direct police to neighborhoods where crime was historically observed \cite{ensign2018runaway, laufer2020compounding}.

\subsection{Neglecting to Consider Multiple Agents} \label{subsec:neglectingtoconsidermultipleagents}

A notable feature of optimization is its strict delineation between decision variables and other (non-decision) variables. Certain parameters are considered to be under the control of the decision-making agent, and other parameters are exogenous and outside the direct control of the agent. This approach lends itself to treating decisions as belonging to a \textit{singular agent}. Even though optimizations are commonly used by institutional bodies and passed between numerous stakeholders ranging from analysts to engineers to executives, an optimization model may not consider that the various constituents of an organization are distinct decision-makers. In this section, we consider how optimization problems may neglect to properly consider the agency and interests of multiple decision-makers, and how such neglects can be avoided or identified. When multiple agents are not taken into consideration, optimizations may only serve the interests of a few people, and additionally may codify or formalize particular structures of management and control. In this section, we consider each of these emergent issues. 

\subsubsection{Objective Represents Narrow Interests} \label{subsubsec:objectiverepresentsnarrowinterests} Optimization is used to make logistical decisions that can impact many people, both inside and outside the decision-making apparatus. For a given optimization, the relative importance of different agents’ utility can be defined in the objective function. When an objective omits or de-values harmful impacts on certain populations, the resulting ``optimal'' decision can be ethically wrongful.

When an objective reflects only certain interests in an organization, the optimal decision might justify exploitative working conditions. Take the toy example of a board of directors deciding how to use a certain amount in earnings. Without union representation on the board, they might decide that their utmost priority is to keep investors happy and therefore pay dividends. With union representation, however, they might adjust their priorities and devote more earnings to employee compensation. The choice of a \textit{goal} or \textit{objective} is highly dependent on the organizational structure.\footnote{We refer readers to \citet{kasy2021fairness}, which puts forward a formal notion of \textit{power} as it relates to designing the objective function.}

\textit{Example: \textbf{Technology Adoption}}. When financial mathematicians began to suggest that optimization could improve portfolio diversification schemes, firms were slow to adopt these schemes. In one academic paper devoted to the reasons why firms did not adopt these measures, a reason for the resistance was the social and political structure of the firm: ``Probably the single most important reason why many financial institutions don't use portfolio optimizers is \textbf{political}. This is because the effective use of an optimizer mandates significant changes in the structure of the organization and the management of the investment process'' \cite[p 32]{michaud1989markowitz}. Firm managers did not want to cede decision-making power to quantitative analysts, suggesting that organizational social structure influences the use of optimization. 

The ``narrow interests'' described so far concern institutional decision-making involving more than one agent. But problems of narrowly defined goals can also have massive impacts on people external to a company or institution, who have no involvement or even awareness of the decision. The notion of \textit{externalities} \cite{ayres1969production, baumol1972taxation, cornes1996theory} (see also \cite{coase2013problem}) helps to describe how unaccounted-for impacts can affect people who might face barriers entering into contracts or fending off negative consequences. People downstream from a river might be negatively impacted by factory pollution and may not even know it. These effects can be the result of negligent or malicious tweaks to an objective function. They highlight the importance of critically evaluating optimizations and their stated goals.

\subsubsection{Decision Model Formalizes Notions of Control} \label{subsubsec:decisionmodelformalizes} The \textit{decision variables} in an optimization are an important indicator for what an agent has control over. In formalizing exactly what decision is assumed to be under the control of an agent or organization, an optimization can codify existing social structures.

\textit{Example: \textbf{Workforce Optimization}}. Industries are fiercely competitive over employee talent, and retaining employees has become an important priority for companies. So-called `workforce optimization' \cite{roy2019enabling, naveh2007workforce, altner2018two} or `staff planning problems' \cite{anderson2001nonstationary, balachandran1982interactive, abernathy1973three} devise strategies to allocate, promote and pay workers in order to maximize retention and efficiency while minimizing costs. Such models make assumptions about a particular type of employee/employer relationship, for example that employees are contractors whose time can be allocated freely; or that layoff recommendations can come from performance metrics rather than human judgments. The clearest examples of decision models formalizing or implicating systems of control are warehouse management and logistics studies that create staffing tools for warehouse managers that balance worker discomfort, worker risk, and efficiency \cite{wruck2017risk, larco2017managing}.\footnote{These tools are further discussed in \cite{gong2011review, de2017warehouse}.}

\subsection{Mislabeling and Mismeasurement} \label{subsec:labelmismeasurement}

The choice of labels and measurements can lead to undesirable outcomes in the use of optimization. Measurements are erroneous when there is a disconnect between a measurement and an intended (or true) value. In this section, we discuss two types of mismeasurement: imprecise measures and biased measures. These issues can arise from a \textit{conceptual error}, whereby proxies or incorrect labels are used instead of a true underlying value; or they can arise from a \textit{measurement error}, whereby instruments or empirical methods do not work as expected. Accuracy is comprised of both precision and the absence of bias in measurement---below, we briefly discuss mislabeling and mismeasurement in terms of these two components.

\subsubsection{Imprecise Labels and Measurements} \label{subsubsec:impreciselabels} 

Imprecise measurements are measurements with a high level of uncertainty or a low level of significance. A measurement can be precise and inaccurate if it tends to detect values to a certain level of precision, but those values are consistently biased. A measurement is imprecise when its measurement exhibits significant random error. 

\textit{Example: \textbf{Portfolio Optimization}}. After Harry Markowitz put forward a method to optimally diversify portfolios in such a way that maximizes expected return \cite{markowitz1991foundations, markowitz2000mean}, practitioners who tried applying the method found that it under-performed human decisions and even simple equal-weighting methods \cite{Jobson70}. In an article aptly titled ``The Markowitz Optimization Enigma: Is `Optimized' Optimal?'' \cite{michaud1989markowitz} it was shown that, systematically, assets with the highest measurement error were also those that had the highest expected return. \citet[p 68]{jorion1992portfolio} described the issue: ``A major drawback with the classical implementation of mean-variance analysis is that it completely ignores the effect of measurement error on optimal portfolio allocations ... optimization systematically overweights the assets with the highest estimation errors, hence overstates the true efficiency of the optimal portfolio.''

\subsubsection{Biased Labels and Measurements} \label{subsubsec:biasedlabels}

Biased labels and measurements are consistently inaccurate in a way that can systematically lead to different outcomes for different subjects of an optimization.

\textit{Example: \textbf{Medical Risk Assessment}}. \citet{obermeyer2019dissecting} study a risk prediction algorithm that is used on 200 million Americans annually to optimally distribute health services. The objective function of the risk assessment algorithm trained the algorithm to predict future health \textit{costs} rather than health \textit{needs}, and due to a historical gap in access across race, the risk scores systematically recommended fewer Black patients receive medical resources. Notably, the paper invokes the algorithm's \textit{objective function} in framing its inquiry into the mechanism of bias: ``An unusual aspect of our dataset is that we observe the algorithm's inputs and outputs as well as its objective function, providing us a unique window into the mechanisms by which bias arises'' \cite[p 3]{obermeyer2019dissecting}.

\subsection{Faulty Solution Steps} \label{subsec:sloppy}

The steps involved in solving an optimization problem, from relaxations to iterative algorithmic solutions, involve assumptions and simplifications that ought to be dealt with carefully and notated precisely. In this section, we discuss two types of neglect in the steps involved with solving optimizations: \textit{faulty use of relaxations} and \textit{faulty use of approximation algorithms}. Our goal in this section is not to critique optimization research, which tends to precisely and clearly characterize the use of approximations, nor to critique all applications of optimization. More often, these areas of neglect can occur in private settings which emphasize speed and minimum viability \cite{ries2009minimum, moogk2012minimum} in product development.

\subsubsection{Faulty Use of Relaxations} \label{subsubsec:sloppyrelaxations}

When a relaxation is used for ease of computation or measurement, it can significantly change the optimization's feasible set and objective function.  As such, it is important to detail when relaxations are being used, and what assumptions are made in the process. When a problem is relaxed, its fundamental structure has changed. The solution set might be much bigger, or smaller, than the true optimization problem, for example. As such, calling a solution to a relaxed problem `optimal' may be misleading without appropriately caveating. When relaxations are purposefully concealed and a solution is described as \textit{optimal}, such a practice is \textit{faulty}. It postures as \textit{legitimate} on false pretenses, even when its recommended choice could be misguided or non-optimal.

\subsubsection{Faulty Use of Approximation Algorithms} \label{subsubsec:sloppyapproximation}

It is common for operations researchers, mathematicians, and computer scientists to focus on a particular category of problem, come up with a way of approximating the solution, and prove that the approximate solution is within a certain error bound compared to the true solution. In optimization, maximization (minimization) problems are approximated with answers that perform below (above) the true optimal, in which case lower (upper) bounds are given. Approximations are applied carelessly when these guaranteed bounds are not given, and the performance guarantee or bound is not specified. Approximation algorithms are applied erroneously if they are derived for a certain type of problem (e.g., convex) and are applied on a different type of problem (e.g., non-convex). Such a behavior can be \textit{faulty} when performance is purposefully over-stated, or uncertainty is concealed.

\subsection{Using “Optimal” as Justification} \label{subsec:usingoptimalasjustification} \label{subsec:just}

The final problem identified in this paper arises when people assume that an \textit{optimal decision} is an \textit{ethically justified decision}. As optimization has entered common parlance, it is used to describe a variety of behaviors and strategies that aim to achieve a variety of ends. Optimization is sometimes invoked without specifying a particular decision or objective---an optimal workforce strategy or an optimal economic policy lack the kind of specificity that is required to understand if a particular strategy is morally or politically desirable. In this section, we discuss the justificatory use of optimization. Particular mechanisms by which optimization can be used to justify action are apparent in language (e.g., when a decision-makers calls their decision \textit{optimal} without further context) and in cases where an optimization problem is retrofit to a decision that has already been made.

\subsubsection{Implicit Subject-Object Relationship} \label{subsubsec:language}

The use of the verb ``optimize,'' when used in the sense we discuss in this paper, implies that there is a \textit{decision} and a \textit{goal}. Sometimes, when people use the word without specifying one of these constitutive components, the context can clarify: ``optimize draft picks" might specify the decision and imply the goal, for example. When the decision and goal are indeterminate and a quantity is claimed to be \textit{optimal}, however, we argue that this description is a category error and its use is a primary suspect for the sort of neglect we are describing. Especially in high-stakes settings, where the goals of a certain decision might be contested, such as admissions, sentencing, and urban policy, using the word \textit{optimal} to describe a particular decision may be a faulty justification or optic maneuver.

\subsubsection{Reverse Engineering} \label{subsubsec:formal}
Even if an optimization model is defined and solved indicating that a particular decision is optimal, such a process does not morally justify making that decision.  This point can be demonstrated by considering two optimizations, nearly identical, except one switches ``minimize" for ``maximize." Both yield \textit{optimal solutions} that have diametrically opposite utility performance. The best possible and worst possible strategies for a given optimization can both be said to be optimal strategies, if the optimization is tweaked. A broader and perhaps deeper point is that \textit{any decision can be framed as the solution to an optimization problem}. By limiting the feasible set enough, and framing the objective in just the right way, any number of optimization problems can yield any decision as its optimal solution. Using optimal as justification, then, does not translate to ethical justification.



\section{Discussion and Suggestions}\label{sec:programme}

Optimization is a paradigm used to aid and support decision-making in potentially high-stakes situations. It offers a systematic way of balancing interests and considerations as objectives or constraints relevant to a particular decision. The required design choices --- including which interests and considerations to model, and how to model them --- are often guided by norms, commitments and priorities. For example, a social media company might use optimization to recommend content in a way that attracts attention and engagement from users. Or, a government might use optimization to set tariffs in a way that promotes growth while satisfying domestic workers and unions. Using optimization in a particular domain requires making choices not just about which factors to include, but also about which factors to omit. Much of the context around a decision is omitted in order to model the decision as an optimization. Accordingly, when an optimization is “solved” and a course of action is selected, this does not suggest that the action is morally or ethically or socially good, \textit{all things considered}. Instead, the so-called `optimal' solution satisfies a very specific set of considerations: namely, those that are specified and introduced through the process of modeling, measuring, computing and applying the paradigm.

In this paper, we put forward a framework for understanding the normative commitments that are contained, either implicitly or explicitly, in the use of mathematical optimization. To do so, we focused on the steps involving abstraction and assumption, many of which are choices that reflect practitioners’ priorities and values. We then attempted to excavate some of the assumptions, areas of neglect, and emergent issues that can arise along the various components of an optimization’s life cycle. 

As researchers are reckoning with the potential for bias, unfairness, inequality, opacity, and other ethical issues in machine learning and dataset creation, a similar line of inquiry is needed to scrutinize the normative (and epistemic) elements of optimization. A number of research questions in this line of inquiry remain, including: which norms, commitments, and priorities might constitute appropriate or legitimate optimization?

These and other considerations around optimization and optimal behavior are open questions that, we hope, will motivate a program of future research. There are a number of directions for work, some of which stem directly from the present paper’s framework.

\textit{Empirical work}. The use of optimization as a paradigm for decision-making requires constructing an abstract model and using it to make decisions in real-world contexts. Empirical studies about optimization’s use and impacts may find particular patterns, conventions, and mechanisms through which optimization causes social harms (or, conversely, benefits). As a notable step, \citet{kulynych2020pots} identify examples of \textit{protective optimization technologies} --- real-world uses of optimization and other techniques that resist and reclaim other, harmful optimization and machine learning systems.\footnote{See also \citet{gurses2018pots}.} Sometimes, simply observing and documenting the components of an optimization (e.g., the objectives) can provide useful empirical insight about an individual or institution’s operations. For example, the fact that a social media company designs features to optimize for user engagement and attention might reveal pressing and legitimate issues about its products.

Available traces and artifacts pointing to the use of optimization are crucial for empirical study and, ultimately, conversations about appropriate use. To that end, \textit{documentation} can play a role in making optimization use transparent and accountable. The work on documenting models \cite{mitchell2019model}, datasets \cite{gebru2021datasheets}, and reinforcement learning models \cite{gilbert2022reward} might hold lessons for optimization. These studies can ultimately enable audits, new standards, best practices, and other mechanisms for accountability.

\textit{Conceptual work}. This paper provides evidence that optimization, as a paradigmatic technical tool, encodes certain normative commitments. Theorizing about whether and when these commitments are justified remains a promising direction for further inquiry. Normative analysis of this sort would benefit from careful consideration of a number of contextual parameters about the use of optimization: By whom? Of what? When? Where? For what end? It would also benefit from an aggregate and system-wide framing on optimization and its political implications. For instance, it seems evident that private corporations tend to optimize for profit, or a proxy thereof. In light of this, how can markets, institutions and processes improve the use of optimization and aim technological tools towards social ends?






\begin{acks}
We thank the Digital Life Initiative (DLI), the AI, Policy and Practice Initiative (AIPP), the Cornell Tech Emma Pierson and Nikhil Garg (CTEN) Lab, and the Cornell University Information Science Colloquium for providing remarks and suggestions. We'd like to acknowledge Nikhil Garg, Yonatan Mintz, Eugene Bagdasaryan, Jon Kleinberg, Karen Levy, and David Williamson for providing invaluable feedback.

The authors are grateful for awards from the US National Science Foundation, NSF CNS-1704527 and the John D. and Catherine T. MacArthur Foundation, which generously supported this work. 
\end{acks}



\balance

\bibliographystyle{ACM-Reference-Format}
\bibliography{sample-base}

\newpage
\appendix


\section{Optimization Definition} \label{app:opt}

Optimization can be defined as a mathematical problem of the form specified in equation (\ref{eq:standardform}), where variables are defined on the (continuous) set of real numbers. 

\begin{eqnarray}
\label{eq:standardform}
\underset{x}{\text{maximize}} & f(x) & \\
\text{subject to} & g_i(x) \leq 0,&  i = 1,2,...,k \nonumber \\
& h_j(x) = 0,&  j = 1,2,...,l  \nonumber
\end{eqnarray}

\ 

The \textbf{decision} is modeled as a vector of $n$ variables $x \in \mathbb{R}^n$. $x$ can only take values that satisfy the $k \geq 0$ inequality constraints and $l\geq0$ equality \textbf{constraints}, $g_i(x) \leq 0$ and $h_j(x) = 0$, respectively. The \textbf{objective function} $f: \mathbb{R}^n \rightarrow \mathbb{R}$ is defined on the \textit{feasible set} dictated by the constraints. The solution is the decision value $x^{OPT}$ which maximizes $f(x)$ out of all feasible values of $x$.


\end{document}